\documentclass[letterpaper,10pt,conference]{ieeeconf}  

\IEEEoverridecommandlockouts                              

\usepackage{mymacros}
\usepackage{times} 
\usepackage{graphicx}

\usepackage[T1]{fontenc}
\usepackage{color}
\usepackage{pifont}

\usepackage{caption} 

\let\labelindent\relax
\usepackage{enumitem}

\title{\LARGE \bf
     Open-Vocabulary Mobile Manipulation Based on \\ 
      Double Relaxed Contrastive Learning with Dense Labeling
}

\author{
    Daichi Yashima, Ryosuke Korekata, and Komei Sugiura
\thanks{
    \small The authors are with Keio University, 3-14-1 Hiyoshi, Kohoku, Yokohama, Kanagawa 223-8522, Japan.
    {\tt\small \{ydaichi1207, rkorekata, komei.sugiura\}@keio.jp}
}
}

\begin{document}

\makeatletter
\let\@oldmaketitle\@maketitle 
\renewcommand{\@maketitle}{\@oldmaketitle 
}
\makeatother

\maketitle

\thispagestyle{empty}
\pagestyle{empty}
\begin{abstract}
Growing labor shortages are increasing the demand for domestic service robots (DSRs) to assist in various settings.
In this study, we develop a DSR that transports everyday objects to specified pieces of furniture based on open-vocabulary instructions.
Our approach focuses on retrieving images of target objects and receptacles from pre-collected images of indoor environments.
For example, given an instruction ``Please get the right red towel hanging on the metal towel rack and put it in the white washing machine on the left,'' the DSR is expected to carry the red towel to the washing machine based on the retrieved images.
This is challenging because the correct images should be retrieved from thousands of collected images, which may include many images of similar towels and appliances.
To address this, we propose RelaX-Former, which learns diverse and robust representations from among positive, unlabeled positive, and negative samples.
We evaluated RelaX-Former on a dataset containing real-world indoor images and human annotated instructions including complex referring expressions.
The experimental results demonstrate that RelaX-Former outperformed existing baseline models across standard image retrieval metrics.
Moreover, we performed physical experiments using a DSR to evaluate the performance of our approach in a zero-shot transfer setting. 
The experiments involved the DSR to carry objects to specific receptacles based on open-vocabulary instructions, achieving an overall success rate of 
75\%.
 \end{abstract}

\vspace{-2.0mm}
\section{Introduction
\label{intro}
}
\vspace{-1mm}
Service robots capable of transporting objects and working alongside humans are becoming increasingly important in various situations such as restaurants, hospitals, and warehouses, especially with rising labor shortages and insufficient workforces in society.
To enhance their functionality, these robots should incorporate natural language understanding capabilities.
This capability is particularly valuable for domestic service robots (DSRs), which can assist elderly people by performing daily tasks.
However, DSRs face significant challenges in identifying the target object or the receptacle from numerous similar objects based on open-vocabulary instructions that include complex referring expressions.

\begin{figure}[t]
    \centering
    \includegraphics[width=\linewidth]{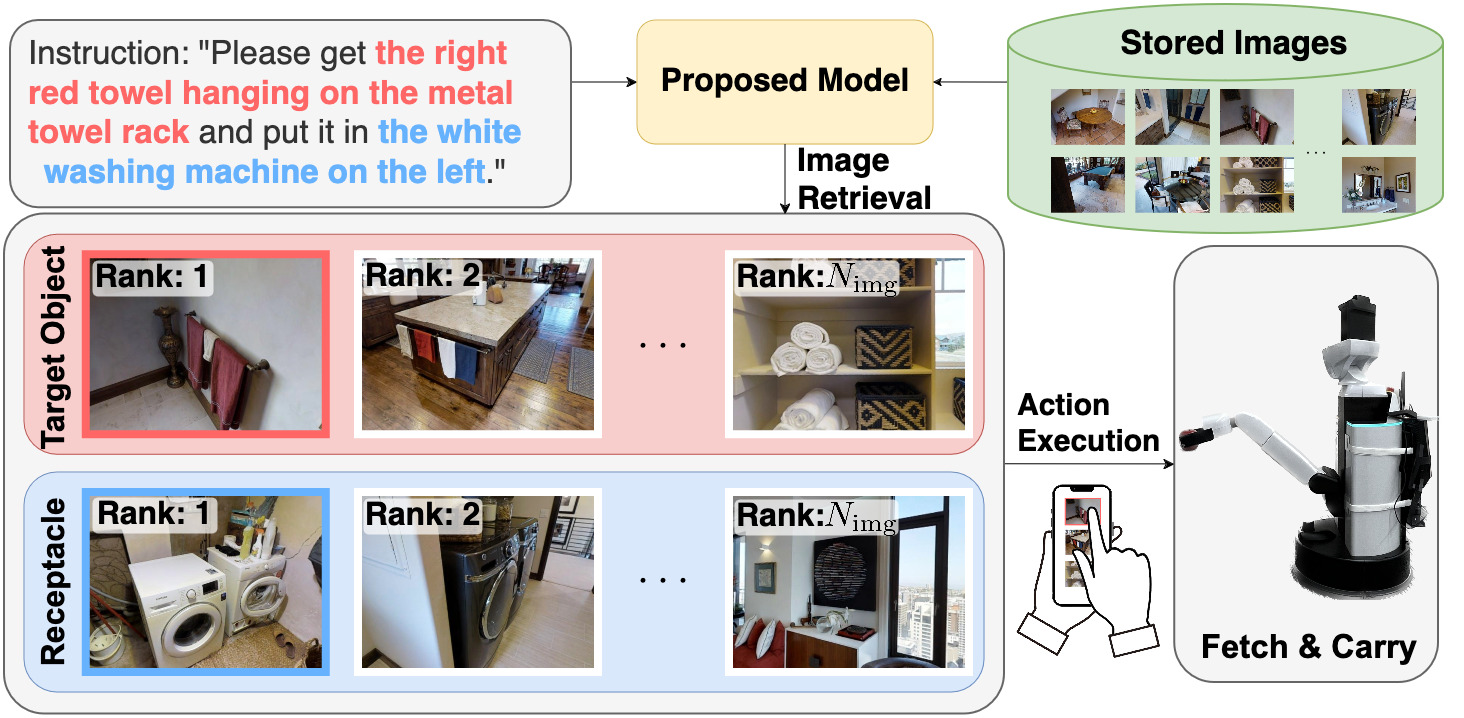}
    \vspace{-5mm}
    \caption{\small Overview of our task. First, the DSR collects images of the environment through pre-exploration. Given an open-vocabulary instruction, it is required to retrieve the red and blue framed images as the target object image and receptacle image, respectively, from the collected images. Subsequently, the DSR carries the target object to the receptacle, based on the user-selected images.}
    \label{fig:eye_catch}
    \vspace{-6mm}
\end{figure}

In this study, we focus on developing a DSR system that transports everyday objects to specific pieces of furniture.
Open-vocabulary instructions are used to retrieve images of target objects and receptacles from a collection of environmental images.
Fig.~\ref{fig:eye_catch} shows a typical scene for this task.
First, the DSR collects images of indoor environments during pre-exploration.
Next, given an instruction such as “Please get the right red towel hanging on the metal towel rack and put it in the white washing machine on the left,” the proposed method outputs two ranked lists: one for the red towel hanging on the metal towel rack and the other for the white washing machine on the left.
The DSR is then tasked with transporting the target object to the receptacle, based on the images selected by users from the two ranked lists.

The primary challenge in this task is to retrieve images from open-vocabulary instructions, which often include complex referring expressions that involve both target objects and receptacles.
Moreover, it is challenging to retrieve the ground-truth image when the DSR has collected many similar images in the environment.
While recent multimodal representation models such as CLIP~\cite{radford21icml} have improved cross-modal retrieval performance, they still have limitations for this task, as discussed in Section~\ref{quantitative_res}. 
Most models use contrastive loss functions such as InfoNCE\cite{oord18arxiv}, where a single phrase is paired with a single positive image, and all other pairs within the batch are considered negative (e.g.,\cite{radford21icml, kaneda24ral, korekata24arxiv}).
This approach can be disadvantageous when the batch contains unlabeled positive samples.
These unlabeled positives are typically treated as negatives, potentially resulting in undesired anti-correlation between samples that should be positively correlated.
The lack of positive annotations often occurs due to the labor-intensive and time-consuming nature of providing detailed labels for all similar instances\footnote{Comprehensively annotating all possible unlabeled positive samples in our training set of 5,814 would require approximately 188,000 h of human labor, assuming 10 s per annotation and checking each instruction against all images.}.

In this paper, we propose 
RelaX-Former, a novel method that leverages unlabeled positive labels and introduces a 
double
relaxed contrastive learning approach to handle unlabeled positive and negative samples, thereby improving the alignment between images and text.
\color{black}
Most conventional methods follow the standard approach of using contrastive loss functions such as InfoNCE, often ignoring the presence of unlabeled positives\cite{radford21icml, lin23wacv}.
In contrast, the proposed method incorporates unlabeled positives by assigning unlabeled positive labels to images similar to the ground-truth image through the Dense Labeler.
The 
Double
Relaxed Contrastive (DRC) loss is then used to handle the relations among positive, unlabeled positive, and negative samples.
\color{black}
By distinguishing between unlabeled positive and negative samples with the Dense Labeler and applying the proposed DRC loss, our approach enables a more diverse and effective learning representation.
Our code is available at this URL\url{}\footnote{https://github.com/keio-smilab24/RelaX-Former}. 

The main contributions of this study are as follows:
\vspace{-1mm}
\begin{itemize}
    \item We propose the Spatial Overlay Grounding (SOG) module, which uses a prompting technique that uses markers on segmented images to obtain features from a multimodal large language model (MLLM). By generating dense captions for multiple regions within an image, we obtain fine-grained, linguistically grounded features that enhance the richness and diversity of visual representations.
    \color{black}
    \item We present the Dense Representation Learning (DRL) module, which employs the Dense Labeler to estimate unlabeled positives to images similar to the positive image and uses the DRC loss to effectively handle the relationship among positive, unlabeled positive, and negative samples.
\end{itemize}

\vspace{-1mm}

\vspace{-1.2mm}
\section{
    Related Work
}
\vspace{-0.8mm}

\subsection{Language-Guided Mobile Manipulation}
There has been widespread research in mobile manipulation tasks guided by natural language instructions\cite{brian22corl, chen23icra, driess23icml}.
For instance, \cite{iocchi15aij, okada19ar, yenamandra23corl} have conducted language-guided mobile manipulation tasks using DSRs within standardized real-world environments, demonstrating the application of natural language instructions in practical scenarios.
While these tasks share similar objectives to ours, the task considered in the present study diverges by employing free-form, open-vocabulary instructions accompanied by referring expressions, as opposed to the use of template-based instructions.
Although \cite{yenamandra23corl} utilizes open-vocabulary instructions, it differs from our study in that it does not incorporate fully free-form instructions.

Various studies have attempted to identify target objects from among similar objects in the environment based on manipulation instructions\cite{korekata23iros, kaneda24ral, sigurdsson23iros}.
Specifically, \cite{kaneda24ral} and \cite{sigurdsson23iros} involve multimodal image retrieval in indoor environments, requiring the output of a ranked list of multiple images that contain the target object described in the instruction.
Our task differs from these in that it handles both the target object and the receptacle within the instruction in the image retrieval setting.
NLMap\cite{chen23icra} employs a CLIP-based approach with open-vocabulary instructions and pre-explored images for target object and receptacle image retrieval. 
While it has a fully automated image retrieval process, our system enhances this by allowing user selection from the top-ranked retrieved images, improving the adaptability to complex scenarios.

\vspace{-1.8mm}
\subsection{Contrastive Learning}
\vspace{-0.8mm}
There have been numerous studies in the field of contrastive learning and self-supervised representation learning\cite{le20ieee}.
Contrastive learning maps similar (positive) samples close together and dissimilar (negative) samples farther apart in the embedding space.
Representative studies address data augmentation (e.g.,\cite{ting20icml, feng22icpr, denize23wacv}), clustering (e.g.,\cite{zheng21iccv}), and loss function designs 
(e.g.,\cite{feng22icpr, gao22neurips})\color{black}.

InfoNCE\cite{oord18arxiv} is widely used as a contrastive loss function (e.g.,\cite{radford21icml, korekata24arxiv, Gao24aaai, wu22iclr})
but its primary limitation is the imbalance between positive and negative samples during training. 
This restricts the diversity and adaptability of the representations by consistently distancing all unpaired samples from the query\cite{le20ieee}.
To address this limitation, softened loss functions such as
\cite{lin23wacv, ge2023icml, feng22icpr} 
have been proposed as more balanced alternatives to InfoNCE. 
ReCo\cite{lin23wacv} relaxes the strict contrast applied to negative samples within the embedding space.
\color{black}
Although this alleviates the asymmetric relationship between positive and negative embeddings, it does not explicitly consider unlabeled positives.
We address this issue by introducing the novel DRC loss, which can handle unlabeled positives and relaxes the spaces of unlabeled positive and negative embeddings, resulting in more refined representations.

\begin{figure*}[t]
    \centering
    \vspace{1.8mm}
    \includegraphics[width=\linewidth]{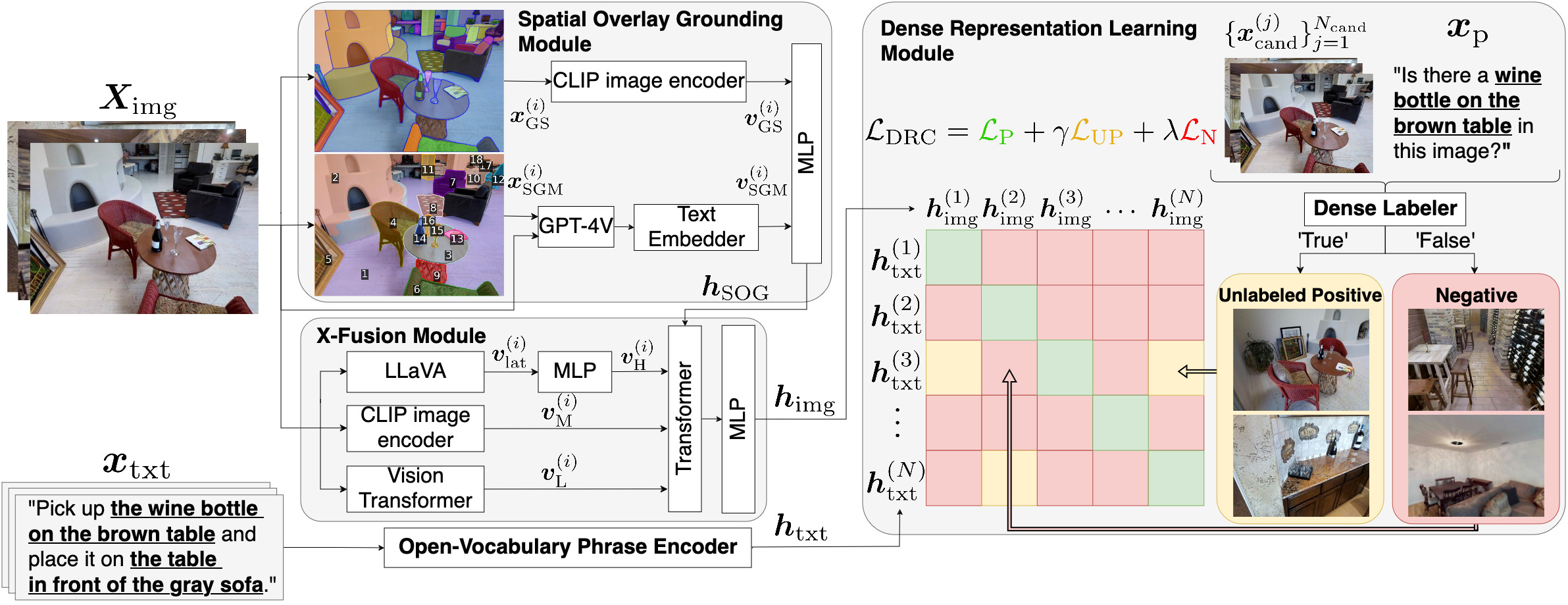}
    \vspace{-4.0mm}
    \caption{\small Architecture of 
    RelaX-Former.
    The proposed architecture consists of four modules: Spatial Overlay Grounding (SOG) module, X-Fusion (XF) module, Dense Representation Learning (DRL) module, and Open-Vocabulary Phrase (OVP) encoder. Here, $N$ denotes the batch size.}
    \label{fig:model}
    \vspace{-7.2mm}
\end{figure*}

\vspace{-2mm}
\section{Problem Statement
}
\vspace{-1mm}

In this study, we focus on the Image Retrieval-based Open-Vocabulary Fetch-and-Carry (IROV-FC) task\cite{korekata24arxiv}.
Fig.~\ref{fig:eye_catch} shows a typical scene associated with this task.
The task procedure is defined as follows:

\begin{enumerate}
    \item The DSR collects images of an indoor environment during pre-exploration. 
    \item Image retrieval: given an instruction such as ``Please get the right red towel hanging on the metal towel rack and put it in the white washing machine on the left,'' two ranked lists of images should be retrieved: one for the target object and the other for the receptacle. 
    Images of the target object and the receptacle should be ranked highly in these lists.
    \item [3)]
    The correct target object and receptacle are selected by the user from among the top-$K$ retrieved images. 
    \item [4)] 
    Action execution: the DSR moves to the location at which the target image was collected, grasps the target object, carries it to the receptacle, and places it there. 
\end{enumerate}


The terminology used in this paper is defined as follows:
\begin{itemize}
    \item \textbf{Target object:} an everyday object identified as the target in the instruction.
    \item \textbf{Receptacle:} a piece of furniture identified as the designated placement area in the instruction.
\end{itemize}

We assume that the images of the indoor environment have already been collected through pre-exploration.
This is reasonable because DSRs are typically used in the same indoor environment for long periods of time\cite{sigurdsson23iros, chen23icra}.
It is also assumed that trajectory generation regarding navigation, object grasping, and object placement is based on heuristic methods.

\vspace{-0.5mm}
\section{Proposed Method
}
\vspace{-0.8mm}

Fig.~\ref{fig:model} shows the structure of our proposed method, RelaX-Former. 
 It consists of four modules: Spatial Overlay Grounding (SOG) module, X-Fusion (XF) module, Dense Representation Learning (DRL) module, and Open-Vocabulary Phrase (OVP) encoder.

The input $\bm{x}$ to RelaX-Former is defined as follows:
\begin{align*}
\bm{x}=\{X_{\text{img}}, \bm{x}_{\text{txt}}, m\},\; X_{\text{img}} = \{\bm{x}_{\text{img}}^{(i)} | \; i = 1,\cdots, N_{\text{img}}\},    
\end{align*}
where $\bm{x}_{\text{txt}} \in \{0, 1\}^{V \times L}$, $m \in \{\langle\text{target}\rangle, \langle\text{receptacle}\rangle\}$  and $\bm{x}_{\text{img}} \in \mathbb{R}^{3 \times W \times H}$ denote a tokenized instruction, an image, and a mode token that indicates the basis for the ranking, respectively.
\color{black}
Here, $V$, $L$, $i$, $N_{\text{img}}$, $W$, and $H$ denote  the vocabulary size, maximum token length, index of each image in the set of collected images to be ranked, number of collected images to be ranked, image width, and image height, respectively.
\vspace{-0.8mm}
\subsection{\textit{Spatial Overlay Grounding Module}}
In the SOG module, visual features are obtained using two parallel streams. One stream applies a multimodal encoder with segmentation masks, while the other employs a MLLM with region-marked images, both leveraging mask information to enhance regional features.
Typical methods obtain visual features either globally or by focusing on a single object or part of an object, which can lead to misinterpretations of the visual context, as shown in Section \ref{ablation}.
On the other hand, our approach uses foundation models for segmentation (e.g.,\cite{kirillov23iccv, zou24neurips}) to isolate objects.
This provides auxiliary information such as contours, shape, and the relative positions between objects, which helps reduce visual errors.

This module takes ${X}_{\text{img}}$ as the input.
We use SAM\cite{kirillov23iccv} and SEEM\cite{zou24neurips} to generate segmentation masks for $\bm{x}_{\text{img}}^{(i)}$.
The masks 
generated by SAM and SEEM 
are overlaid on $\bm{x}_{\text{img}}^{(i)}$ to obtain $\bm{x}_{\text{GS}}^{(i)} \in \mathbb{R}^{3 \times W \times H}$ and $\bm{x}_{\text{SS}}^{(i)} \in \mathbb{R}^{3 \times W \times H}$, respectively.
Next, visual features $\bm{v}_{\text{GS}}^{(i)} \in \mathbb{R}^{d_\mathrm{GS}}$ are obtained using a pre-trained CLIP image encoder (ViT-L/14)\cite{radford21icml} on $\bm{x}_{\text{GS}}^{(i)}$.
Subsequently, 
as shown in Fig.~\ref{fig:model},
we obtain $\bm{x}_{\text{SGM}}^{(i)} \in \mathbb{R}^{3 \times W \times H}$ by assigning numerical marks to each segmentation mask in $\bm{x}_{\text{SS}}^{(i)}$.
This process consists of two steps: determining the mark location and assigning numerical labels.
For mark location determination, we process masks from the smallest to largest area, determining suitable positions while excluding regions covered by previously processed masks.
The final numerical labels are then assigned to these locations in descending order following \cite{yang2023arxiv}.
\color{black}

An image description is then obtained using an MLLM based on three inputs: $\bm{x}_{\text{img}}^{(i)}$, $\bm{x}_{\text{SGM}}^{(i)}$, and a text prompt that instructs the description of the structured relationship of the room.
The image description is embedded with text-embedding-large-3, and the resulting embedding is processed through a multi-layer perceptron (MLP) to obtain $\bm{v}_{\text{SGM}}^{(i)} \in \mathbb{R}^{d_{\mathrm{SGM}}}$.
To enhance the generation of image descriptions by MLLMs, we adopt a dual-input strategy that uses both $\bm{x}_{\text{img}}^{(i)}$ and $\bm{x}_{\text{SGM}}^{(i)}$ for inputs to GPT-4V.
This addresses the issue of the marks overlapping and potentially hiding essential image details.
While also preventing misinterpretation of the masked colors as those from the actual scene by providing the raw image for correct color recognition.
Finally, the output $\bm{h}_{\text{SOG}} \in \mathbb{R}^{d_{\mathrm{SOG}}}$ is obtained by concatenating $\bm{v}_{\text{GS}}^{(i)}$ and $\bm{v}_{\text{SGM}}^{(i)}$ and feeding it into a MLP.

\vspace{-0.8mm}
\subsection{X-Fusion Module}
In the XF module, we comprehensively obtained three types of visual embeddings from pre-trained visual encoders(e.g., \cite{dosovitskiy2021iclr}), multimodal encoders(e.g., \cite{radford21icml}), and MLLMs that provide latent features (e.g., \cite{liu2024neurips}).
Visual encoders capture edges, textures, and shapes, but lack semantic and spatial understanding. 
Multimodal encoders provide semantically aligned embeddings that connect visual and textual data, though they struggle with spatial detail. 
In contrast, MLLMs that provide latent features can obtain structural features that directly represent spatial expressions and complex referring relations through embeddings.
Using these latent features are advantageous over text-based outputs because they eliminate the need to pass through a tokenizer, allowing for straightforward utilization of the features.
Thus, we use these three types of embeddings in parallel to leverage their complementary strengths.
Furthermore, these three specific feature types were chosen based on the results of preliminary experiments.
\color{black}

The XF module takes ${X}_{\text{img}}$ as input.
First, we derive the visual features $\bm{v}_{\text{L}}^{(i)} \in \mathbb{R}^{d_{\mathrm{L}}}$ from a pre-trained visual encoder (ViT).
Next, $\bm{v}_{\text{M}}^{(i)} \in \mathbb{R}^{d_\mathrm{M}}$ is obtained by the pre-trained CLIP image encoder.
The latent features $\bm{v}_{\text{lat}}^{(i)} \in \mathbb{R}^{d_\mathrm{lat}}$ are acquired using an MLLM (LLaVA-v1.6-mistral-7b) from $\bm{x}_{\text{img}}^{(i)}$ and a prompt that instructs the MLLM to describe $\bm{x}_{\text{img}}^{(i)}$.
We feed $\bm{v}_{\text{lat}}^{(i)} $ into a MLP to obtain $\bm{v}_{\text{H}}^{(i)} \in \mathbb{R}^{d_\mathrm{H}}$.
Finally, the comprehensive visual features $\bm{h}_{\text{img}} \in \mathbb{R}^{d_\mathrm{img}}$ are obtained as follows:
\begin{align*}
        \bm{h}_{\text{img}} = \text{MLP}\left(\text{Transformer}\left(\left [\bm{v}_{\text{L}};\bm{v}_{\text{M}};\bm{v}_{\text{H}};\bm{h}_{\text{SOG}}\right]\right)\right),
\end{align*}
where $\text{Transformer}(\cdot)$ denote the transformer encoder.

\vspace{-0.8mm}
\begin{table*}[t]
    \vspace{1.2mm}
    \centering
    \caption{Quantitative comparison between RelaX-Former and baseline methods on the test sets. The best score for each metric is presented in \textbf{bold}.
    Recall scores were calculated individually for each test environment and the average of these scores was reported.
    \color{black}
    }
    \vspace{-1.2mm}
    \resizebox{\textwidth}{!}{%
    \begin{tabular}{@{}clcccccc@{}}
    \toprule
    \multicolumn{1}{l}{\multirow{2}{*}{}} & \multicolumn{1}{l}{\multirow{2}{*}{Method}} & \multicolumn{3}{c}{HM3D-FC (unseen)} & \multicolumn{3}{c}{MP3D-FC (unseen)} \\ \cmidrule(lr){3-5} \cmidrule(lr){6-8}
    \multicolumn{2}{c}{} & \multicolumn{1}{c}{R@5$\uparrow$\;[\%]} & \multicolumn{1}{c}{R@10$\uparrow$\;[\%]} & \multicolumn{1}{c}{R@20$\uparrow$\;[\%]} & \multicolumn{1}{c}{R@5$\uparrow$\;[\%]} & \multicolumn{1}{c}{R@10$\uparrow$\;[\%]} & \multicolumn{1}{c}{R@20$\uparrow$\;[\%]} \\ \midrule
    (i) & NLMap \cite{chen23icra} (rep.)\footnotemark & 14.7 & 27.9 & 53.2 & 12.2 & 27.1 & 63.8 \\
    (ii) & MultiRankIt \cite{kaneda24ral}  & 28.7 \scriptsize{$\pm$ 3.4} & 48.3 \scriptsize{$\pm$ 3.4} & 73.3 \scriptsize{$\pm$ 2.6} & 35.7 \scriptsize{$\pm$ 9.9} & 51.7 \scriptsize{$\pm$ 8.9} & 72.7 \scriptsize{$\pm$ 3.3} \\
    (iii) & $\text{DM}^{2}\text{RM}$ \cite{korekata24arxiv} & 47.8 \scriptsize{$\pm$ 1.2} & 67.1 \scriptsize{$\pm$ 2.4} & 87.0 \scriptsize{$\pm$ 1.1} & 49.6 \scriptsize{$\pm$ 0.7} & 64.1 \scriptsize{$\pm$ 3.6} & 78.5 \scriptsize{$\pm$ 0.5} \\ \midrule
    (iv) & RelaX-Former (ours) & \textbf{55.4} \scriptsize{$\pm$ 0.5} & \textbf{76.3} \scriptsize{$\pm$ 0.9} & \textbf{91.6} \scriptsize{$\pm$ 0.9} & \textbf{57.0} \scriptsize{$\pm$ 1.1} & \textbf{72.4} \scriptsize{$\pm$ 0.7} & \textbf{82.5} \scriptsize{$\pm$ 0.8} \\
    \bottomrule
    \end{tabular}%
    }
    \label{tab:result}
\vspace{-4mm}
\end{table*}
\subsection{Open-Vocabulary Phrase Encoder}
The OVP encoder, following \cite{korekata24arxiv}, efficiently processes open-vocabulary instructions for both target and receptacle modes within a single model framework. 
It takes $\bm{x}_{\text{txt}}$ and $m$ as the input.
An LLM identifies the phrases of either the target object or the receptacle based on the mode. It then generates standardized instructions, while a parser extracts noun phrases.
Note that a standardized instruction in this case is a paraphrased instruction with a format for a typical fetch-and-carry task such as ``Carry A to B.''\color{black}
The encoder combines CLIP-processed text features from the original instruction, standardized instruction, and mode-specific phrases. These features, along with transformer-encoded outputs of the extracted phrases, are processed through a multi-layer perceptron to obtain the final text representation $\bm{h}_{\text{txt}} \in \mathbb{R}^{d_{\mathrm{{txt}}}}$.

\vspace{-0.8mm}
\subsection{Dense Representation Learning Module}
The DRL module leverages the novel DRC loss to optimize among positive, unlabeled positive, and negative samples through a relaxed contrastive approach.
Recent multimodal pre-training methods (e.g.,\cite{radford21icml}) primarily use contrastive loss functions such as InfoNCE\cite{oord18arxiv}.
With InfoNCE, the model is optimized to maximize the similarity between the positive samples and minimize the similarity between the negative samples, targeting similarity scores closer to $1$ and $-1$, respectively.
However, in scenarios where the dataset contains similar images, this method results in those similar images being treated as negative samples when they should actually be considered positive or partially positive.
The situation where only one annotation per ground truth is provided is due to various constraints, such as the labor-intensive and time-consuming costs associated with providing annotations as discussed in Section~\ref{intro}. 
Therefore, the application of annotating unlabeled positives and a loss function that can appropriately handle these cases is crucial.

To address this issue, we propose the following DRC loss:
$\mathcal{L}_{\mathrm{DRC}} = \mathcal{L}_{\text{P}} + \gamma\mathcal{L}_{\text{UP}} + \lambda\mathcal{L}_{\text{N}},$
where $\gamma$ and $\lambda$ are hyperparameters that control the weights of the unlabeled positive sample loss and the negative sample loss, respectively.
\color{black}
The components constituting $\mathcal{L}_{\mathrm{DRC}}$ represent the losses for positive, unlabeled positive, and negative samples, respectively, which are defined as follows:
\begin{align*}
    \mathcal{L}_{\text{P}} &= \sum_{i} \left( 1 - \mathrm{sim}\left( \bm{h}_{\text{txt}}^{(i)}, \bm{h}_{\text{img}}^{(i)} \right) \right)^2, \\
    \mathcal{L}_{\text{UP}} &= \sum_{(i,j) \in \mathcal{S}} \max\left( \alpha - \mathrm{sim}\left( \bm{h}_{\text{txt}}^{(i)}, \bm{h}_{\text{img}}^{(j)} \right), 0\right)^2, \\
    \mathcal{L}_{\text{N}} &= \sum_{\substack{(i,j) \notin \mathcal{S}}}  \max\left( \mathrm{sim}\left( \bm{h}_{\text{txt}}^{(i)}, \bm{h}_{\text{img}}^{(j)} \right), 0 \right)^2,
\end{align*}
where sim$(\cdot, \cdot)$, $\mathcal{S}$, and $\alpha$ denote the similarity score, the set of indices corresponding to unlabeled positive samples, and a margin parameter that sets a threshold for the similarities of unlabeled positive samples, respectively.
We use cosine similarity for $\mathrm{sim}(\bm{h}_{\mathrm{txt}}^{(i)}, \bm{h}_{\mathrm{img}}^{(j)})$.


We introduce the Dense Labeler, which uses an MLLM (LLaVA-v1.6-mistral-7b) to label the unlabeled positives.
The input is a text prompt and a set of candidate images.
The text prompt describes whether the target object or receptacle can be seen in the image, depending on the mode.
First, all scores $  -1 \le \text{sim}(\bm{h}_{\text{txt}}^{(i)} , \bm{h}_{\text{img}}^{(j)}) \le 1$ are obtained using existing pre-trained models that have been successfully applied in image retrieval tasks (e.g., \cite{radford21icml, kaneda24ral}).
\color{black}  
After obtaining the score for $\bm{x}_{\text{txt}}^{(i)}$, the top $N_{\text{cand}}$ images are selected as candidates and input into the MLLM. 
If the output text includes `True', the image is classified as an unlabeled positive and the indices $(i, j)$ are included in $\mathcal{S}$. 

With the obtained $\mathcal{S}$, $\mathcal{L}_{\text{SP}}$ penalizes the model for samples $(i, j) \in \mathcal{S}$ where $\text{sim}(\bm{h}_{\text{txt}}^{(i)},\bm{h}_{\text{img}}^{(j)})$ is less than $\alpha$.
The $\text{max}$ function allows only those samples with a similarity score of less than $\alpha$ to contribute to the loss, thereby relaxing the contrastiveness by disregarding samples that have similarity scores greater than $\alpha$.
The parameters $\alpha$ and $\gamma$ represent the leniency of the unlabeled positive labels and the reliance on the Dense Labeler, respectively.
The same approach is applied to $\mathcal{L}_{\text{N}}$, where the model is penalized for negative samples $(i, j) \notin\mathcal{S}$ that have  similarity scores higher than 0 by summing the squared similarities. 
By incorporating these components, the DRC loss balances the contribution of positive, unlabeled positive, and negative samples, leading to a more diverse and effective training process.

The output of RelaX-Former during inference is the ranked list of $X_{\mathrm{img}}$ arranged in descending order based on the similarity score $\mathrm{sim}(\bm{h}_{\mathrm{txt}}, \bm{h}_{\mathrm{img}}^{(i)})$.
Two ranked image lists, $\hat{Y}_{\mathrm{targ}}$ and $\hat{Y}_{\mathrm{rec}}$, for the target object and receptacle, respectively, are obtained through a total of two inferences by changing the mode.
\color{black}

\vspace{-1.8mm}
\section{
    Experiments
    \label{sim_exp}
}

\vspace{-1.5mm}

\subsection{
    Dataset
}
\vspace{-0.5mm}

We used the LTRRIE-FC dataset\cite{korekata24arxiv} for the IROV-FC task.
The LTRRIE-FC dataset was constructed from images collected from HM3D\cite{ramakrishnan21neurips, yadav23cvpr} and MP3D\cite{anderson18cvpr, chang173dv}, featuring various everyday environments, with natural language instructions annotated by humans.
Each instruction includes a target object and a receptacle, detailing the task of transporting the target object to the receptacle (e.g., ``Pick up the green vase on the wash basin and put it on the counter-top table in the dining room.'').

HM3D and MP3D are standard datasets in research involving everyday environments, such as navigation and scene understanding \cite{qi20cvpr, sigurdsson23iros}.
To the best of our knowledge, there is no standard dataset that includes human-annotated instructions for fetch-and-carry tasks in both the HM3D and MP3D environments. 
Therefore, we used the LTRRIE-FC dataset.
The dataset consists of 6,581 instructions and 7,148 images collected from 774 environments.
The vocabulary size, total number of words, and average sentence length were 2,491, 103,263, and 15.69, respectively.
In the LTRRIE-FC dataset, the training, validation, and test sets contained 5,814, 354, and 413 samples, respectively.
These sets covered 690, 42, and 42 environments, respectively, with no duplication of environments.
We used the train set to train the baseline and proposed models, the validation set to tune their hyperparameters, and the test sets to evaluate the models.

\footnotetext{The results are based on our reproduction of NLMap, as the original code is not publicly available.}
\begin{figure*}[t]
    \vspace{1.2mm}
    \centering
    \includegraphics[width=\linewidth]{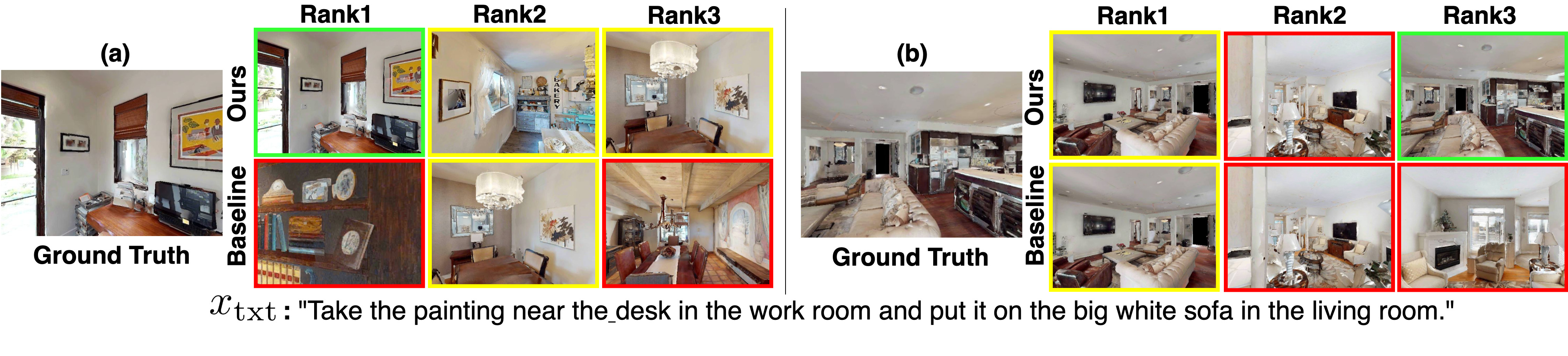}
    \vspace{-6mm}
    \caption{\small Qualitative results of RelaX-Former and the most competitive baseline method \cite{korekata24arxiv} from the HM3D-FC test for $\bm{x}_{\text{txt}}$. The ground-truth image and the top-3 retrieved images are shown for each mode. (a) Target mode and (b) receptacle mode. Positive, unlabeled positive, and negative labels are colored in \textcolor[rgb]{0.2,0.8,0}{green}, \textcolor[RGB]{224, 155, 4}{yellow}, and \textcolor[rgb]{1,0,0}{red}, respectively.
}
    \label{fig:result}
    \vspace{-5.0mm}
\end{figure*}
\vspace{-1.8mm}
\subsection{Parameter Settings}
The AdamW optimizer ($\beta_1 = 0.9$, $\beta_2 = 0.98$) was adopted for training with a learning rate of $1.0 \times 10^{-4}$ and, batch size of $N = 128$.
We used $\alpha = 0.7$ across all the experiments in this study.
The number of candidate images selected for input into the MLLM, denoted as $N_{\text{cand}}$, was set to $20$.


Our model had approximately 201M trainable parameters and a total of 329G multiply-add operations.
We trained our model using a GeForce RTX4090 with 24 GB of GPU memory and an Intel Core i9-13900KF with 64 GB of RAM. 
The training process, which consisted of $20$ epochs, took approximately 3 h.
The inference time for computing the similarity between a single instruction and 100 images was approximately 79 ms.
All visual features required for training and inference, including those from ViT, CLIP image encoder, LLaVA, GPT-4V, SAM, and SEEM, were obtained offline prior to training and the inference process, as these models remain frozen and do not need to be recomputed for each training or inference pass.
 The evaluation on the test sets was based on the model achieving the maximum value of $\text{recall}$@$10$ on the validation set.

\vspace{-1.0mm}
\subsection{Quantitative Results}
\label{quantitative_res}
Table~\ref{tab:result} presents the quantitative results for the performance of the baseline and proposed methods on the HM3D-FC and MP3D-FC test sets. 
The average and standard deviation values from five trials are included.
We used  NLMap\cite{chen23icra}, MultiRankIt\cite{kaneda24ral}, and DM$^2$RM\cite{korekata24arxiv} as the baseline methods.
NLMap was chosen for its ability to retrieve images from a pre-explored set using a method based on CLIP\cite{radford21icml}, which is a representative method for zero-shot image retrieval tasks.
Additionally, we selected MultiRankIt and DM$^2$RM following their successful application to tasks closely related to the IROV-FC task.
The scores presented for NLMap were based on a single trial because a pre-trained frozen model provides consistent results across multiple trials.
Furthermore, we trained MultiRankIt on target objects and receptacles separately and measured the average scores of the two trained models, as this approach cannot handle both modes within a single model.
We used $\text{recall}$@$K\; (K=5, 10, 20)$ as evaluation metrics, with $\text{recall}$@$10$ as the primary metric
because it is a standard metric for image retrieval tasks\cite{cao22ijcai}.
Recall scores were calculated individually for each test environment, and the average of these scores was reported.

Table~\ref{tab:result} indicates that the proposed method (iv) achieved the highest $\text{recall@}10$ scores of 76.3\% on HM3D-FC and 72.4\% on MP3D-FC. This represents improvements of 48.4 points and 45.3 points over (i), 28.0 points and 20.7 points over (ii), and 9.2 points and 8.3 points over (iii), respectively.
Moreover, the proposed method (iv) outperformed the baseline methods across other recall metrics.
The performance differences were statistically significant ($p < 0.01$).
We believe this was because our proposed method handled unlabeled positives effectively.

\subsection{Qualitative Results  \& Discussions}

Fig.~\ref{fig:result} shows successful examples from the LTRRIE-FC dataset using the proposed method and compared them with the most competitive baseline method\cite{korekata24arxiv}.
For each panel, a ground-truth image and the top-3 retrieved images are shown for each mode.
Each retrieved image has a border color indicating its label: \textcolor[rgb]{0.2,0.8,0}{green} for positive, \textcolor[RGB]{224, 155, 4}{yellow} for unlabeled positive, and \textcolor[rgb]{1,0,0}{red} for negative.

\begin{table*}[t]
    \vspace{1.2mm}
    \centering
    \caption{Ablation studies on the test sets. The highest values for each metric are highlighted in \textbf{bold}.
    Recall scores were calculated individually for each test environment, and the average of these scores was reported. }
    \resizebox{\textwidth}{!}{%
    \begin{tabular}{@{}clcccccc@{}}
        \toprule
        \multicolumn{1}{c}{} & \multicolumn{1}{l}{\multirow{2}{*}{Model}} & \multicolumn{3}{c}{HM3D-FC (unseen)} & \multicolumn{3}{c}{MP3D-FC (unseen)} \\ \cmidrule(lr){3-5} \cmidrule(lr){6-8}
        \multicolumn{2}{c}{} & \multicolumn{1}{c}{R@5$\uparrow$\;[\%]} & \multicolumn{1}{c}{R@10$\uparrow$\;[\%]} & \multicolumn{1}{c}{R@20$\uparrow$\;[\%]} & \multicolumn{1}{c}{R@5$\uparrow$\;[\%]} & \multicolumn{1}{c}{R@10$\uparrow$\;[\%]} & \multicolumn{1}{c}{R@20$\uparrow$\;[\%]} \\ \midrule
        (a) & RelaX-Former (full) & \textbf{55.4} \scriptsize{$\pm$ 0.5} & \textbf{76.3} \scriptsize{$\pm$ 0.9} & {91.6} \scriptsize{$\pm$ 0.9} & \textbf{57.0} \scriptsize{$\pm$ 1.1} & \textbf{72.4} \scriptsize{$\pm$ 0.7} & \textbf{82.5} \scriptsize{$\pm$ 0.8} \\ \midrule
        (b) & w/o $\bm{v}_{\text{GS}}^{(i)}$ & 51.6 \scriptsize{$\pm$ 1.2} & 73.5 \scriptsize{$\pm$ 1.1} & \textbf{91.8} \scriptsize{$\pm$ 0.6} & 56.6 \scriptsize{$\pm$ 2.3} & 70.8 \scriptsize{$\pm$ 0.9} & 80.6 \scriptsize{$\pm$ 0.6} \\
        (c) & w/o $\bm{v}_{\text{SGM}}^{(i)}$ & 51.0 \scriptsize{$\pm$ 1.7} & 73.8 \scriptsize{$\pm$ 1.6} & 91.5 \scriptsize{$\pm$ 1.0} & 55.8 \scriptsize{$\pm$ 1.7} & 70.5 \scriptsize{$\pm$ 0.9} & 81.0 \scriptsize{$\pm$ 1.1} \\
        \midrule
        
        (d)\color{black} & w/o $\bm{v}_{\text{L}}^{(i)}$ & 54.2 \scriptsize{$\pm$ 2.1} & 74.8 \scriptsize{$\pm$ 0.7} & 89.9 \scriptsize{$\pm$ 0.8} & 55.9 \scriptsize{$\pm$ 1.7} & 71.1 \scriptsize{$\pm$ 0.7} & 81.6 \scriptsize{$\pm$ 1.2} \\
        (e)\color{black} & w/o $\bm{v}_{\text{M}}^{(i)}$ & 50.7 \scriptsize{$\pm$ 1.6} & 72.3 \scriptsize{$\pm$ 1.4} & 90.9 \scriptsize{$\pm$ 0.4} & 54.3 \scriptsize{$\pm$ 1.1} & 69.9 \scriptsize{$\pm$ 1.3} & 82.0 \scriptsize{$\pm$ 1.6} \\
        (f)\color{black} & w/o $\bm{v}_{\text{H}}^{(i)}$ & 51.2 \scriptsize{$\pm$ 2.0} & 72.9 \scriptsize{$\pm$ 3.0} & 90.2 \scriptsize{$\pm$ 0.8} & 55.8 \scriptsize{$\pm$ 1.4} & 70.6 \scriptsize{$\pm$ 1.1} & 81.6 \scriptsize{$\pm$ 1.3} \\
        \midrule
        (g)\color{black} & w/ InfoNCE\cite{oord18arxiv} & 48.8 \scriptsize{$\pm$ 0.9} & 70.9 \scriptsize{$\pm$ 0.5} & 91.5 \scriptsize{$\pm$ 0.5} & 54.8 \scriptsize{$\pm$ 0.8} & 69.5 \scriptsize{$\pm$ 0.8} & 81.8 \scriptsize{$\pm$ 1.1} \\
        (h) & w/ ReCo\cite{lin23wacv} & 52.5 \scriptsize{$\pm$ 1.4} & 73.5 \scriptsize{$\pm$ 1.1} & 91.4 \scriptsize{$\pm$ 0.7} & 55.4 \scriptsize{$\pm$ 0.7} & 69.1 \scriptsize{$\pm$ 1.3} & 80.9 \scriptsize{$\pm$ 1.3} \\ 
        (i)\color{black} & unlabeled positives as positive & 52.7 \scriptsize{$\pm$ 1.6} & 73.7 \scriptsize{$\pm$ 0.9} & 91.0 \scriptsize{$\pm$ 1.0} & 55.7 \scriptsize{$\pm$ 1.4} & 70.4 \scriptsize{$\pm$ 0.9} & 82.3 \scriptsize{$\pm$ 0.6} \\
        (j)\color{black} & DRC, $\alpha = 0.999999 $ & 51.0 \scriptsize{$\pm$ 1.1} & 73.1 \scriptsize{$\pm$ 0.4} & 90.2 \scriptsize{$\pm$ 0.5} & 55.0 \scriptsize{$\pm$ 1.3} & 70.9 \scriptsize{$\pm$ 0.9} & 80.6 \scriptsize{$\pm$ 0.4} \\
        \bottomrule
    \end{tabular} %
    }
    \label{tab:ablation}
    \vspace{-4mm}
\end{table*}
Figs.~\ref{fig:result} (a) and (b) show an example from the HM3D-FC test set where $\bm{x}_{\text{txt}}$ was ``Take the painting near the desk in the work room and put it on the big white sofa in the living room.''
In (a), the baseline method failed to retrieve the ground-truth image, while the proposed method successfully retrieved the ground-truth image as the top result, with the second and third ranks also including unlabeled positives.
In (b), the baseline method did not include the ground-truth image in the top-3 retrieved images, instead ranking it 9th.
Conversely, the proposed method retrieved the correct image in rank 3 and also retrieved an image with a similar receptacle at rank 1.
This demonstrates that the DRC loss effectively improved the model's ability to retrieve positive and unlabeled positive images.

We further performed detailed error analysis to better understand the limitations.
We defined samples with a rank less than 10 as failure samples.
There were a total of 72 failure samples with 15 and 57 from HM3D-FC and MP3D-FC, respectively.
We performed error analysis on 20 randomly sampled failed samples, 10 from each test set.
Each sample was analyzed with the top-5 images for each mode.
The main error for our model occurred in scenarios where similar images were among the top-ranked images, while the ground-truth image was not ranked within the top-20.
Therefore, in future studies, we plan to develop a Dense Labeler that can assign accurate positive labels to similar images and incorporate a reranking model during inference.
\vspace{-1.8mm}
\subsection{Ablation Studies}
\label{ablation}
Table~\ref{tab:ablation} presents the results of our ablation studies.
The ablation conditions were as follows:

\textbf{SOG ablation:}\;
We conducted experiments by selectively removing streams within the SOG module to investigate their respective effect on performance.
The results show a decrease in the $\text{recall@}10$ for models (b) and (c) compared with model (a) on the HM3D-FC and MP3D-FC test sets.
These results imply that both streams within the SOG module substantially contribute to overall model performance, with the results indicating that segmenting images to isolate specific features enhances the image grounding capabilities of MLLMs and image encoders.

\textbf{XF ablation:}\;
Similarly,
while keeping the Transformer and MLP components we removed the features that were obtained and input into the Transformer within the XF module
to assess its contribution.
The results show a decrease in the $\text{recall@}10$ for models (d), (e), and (f) compared with model (a) on the HM3D-FC and MP3D-FC test sets.
These results suggest that the presence of the XF module which integrates latent features from the MLLM alongside standard visual and multimodal encoders, enhances the model's ability to capture complex multimodal relationships.

\begin{table}[t]

    \vspace{-1.2mm} \caption{ \small Recall@$10$ scores on the test sets across different models and loss functions. The highest scores between the two loss functions for each metric are highlighted in \textbf{bold}.}
    \label{tab:loss_ablation}
    \centering
    \normalsize
    \begin{tabular}{l@{}ll@{}cc}
       \toprule
       \vspace{0.1mm}
        [\%] & \;Model & Loss & \multicolumn{1}{c}{HM3D-FC} & \multicolumn{1}{c}{MP3D-FC} \\
        \midrule
        (i) & \;\multirow{2}{*}{MultiRankIt \cite{kaneda24ral}} & InfoNCE & 48.3 \scriptsize{$\pm$ 3.4} & 51.7 \scriptsize{$\pm$ 8.9}  \\
         (ii)   &  & DRC &  \textbf{57.6} \scriptsize{$\pm$ 0.9} &\textbf{58.3} \scriptsize{$\pm$ 1.9}  \\
        \midrule
        (iii) & \;\multirow{2}{*}{DM$^{2}$RM \cite{korekata24arxiv}} & InfoNCE &  67.1 \scriptsize{$\pm$ 2.4} & 64.1 \scriptsize{$\pm$ 1.7} \\
        (iv)  & & DRC     &  \textbf{69.0} \scriptsize{$\pm$ 2.1} &\textbf{66.8} \scriptsize{$\pm$ 1.1} \\
        \bottomrule
    \end{tabular}
    \vspace{-6mm}
\end{table}


\textbf{Comparison with classic contrastive learning variants:}\;
We compared our proposed loss function with four classic contrastive learning approaches: using InfoNCE\cite{oord18arxiv} (g), ReCo\cite{lin23wacv} (h), treating all unlabeled positive samples as positive (i), and setting $\alpha$ to a soft target value (j). 
In all cases, we observed decreased performance for all evaluation metrics on both the HM3D-FC and MP3D-FC test sets compared with model (a). 
These results demonstrate that our method's selective approach to managing unlabeled positives and the incorporation of the DRL module contribute significantly to improved retrieval performance.

\textbf{Baseline comparisons with different loss functions:}\;
We conducted further experiments on using different loss functions across baseline methods.
Specifically, we compared our proposed DRC loss with the InfoNCE loss across two baselines: MultiRankIt and DM$^2$RM.
Note that we did not include NLMap in the experiment because it is a CLIP-based training free-method.
Table~\ref{tab:loss_ablation} shows the results of our experiment.
In both cases, we observed increased performnace in all evaluation metrics on the HM3D-FC and MP3D-FC test sets compared to the original baseline methods.
These consistent improvements across both architectures demonstrate that our proposed DRC loss effectively addresses the limitations of InfoNCE by better handling unlabeled positives in the training data.

\vspace{-2.5mm}
\section{
    Physical Experiments
}
\vspace{-0.5mm}
We also validated the proposed method using a DSR in zero-shot, real-world experiments.
The DSR executed fetch-and-carry tasks based on open-vocabulary instructions given by the users.
These experiments involved objects that had not been seen during the training phase.

\vspace{-1.8mm}

\subsection{Settings}
The experimental setup closely follows that of DM$^2$RM~\cite{korekata24arxiv}.
The test area measured $4.0 \times 6.0$ m$^2$ and contained nine pieces of furniture arranged in a specific layout.
In the experiments, we used Toyota Motor Corporation's Human Support Robot, which has been the standard platform for RoboCup@Home since 2017~\cite{iocchi15aij}. 
We used 30 everyday objects from the YCB object set~\cite{calli15ram}.
The experiments consisted of 
100 
episodes across 20 different environmental setups, with five episodes per setup. 
In each environmental setup, 15 to 20 objects were randomly selected and placed at arbitrary locations on various pieces of furniture.
There were approximately 400 different images throughout the physical experiment.

\vspace{-1.8mm}
\subsection{Implementation}
The DSR collected RGB-D images of the environment using predetermined waypoints.
The users then randomly selected one target object and one receptacle and provided a unique open-vocabulary fetch-and-carry instruction for each episode from the given environmental setup.
The language instructions were provided by three human annotators from different countries with different cultural backgrounds.
Instructions typically contained referring expressions and involved tasks such as picking up and carrying objects to specific pieces of furniture.
After receiving instructions from the users, the DSR retrieved images from the pre-collected images using the proposed model in the zero-shot setting.
The users then selected an appropriate image for the target object and the receptacle from the 
candidates displayed on a web interface.
If there was no appropriate image of the target object among the top-$K$
(=10)
 candidates, the task was considered a failure, and the DSR did not perform any further steps in the episode.

Subsequently, the DSR navigated to the location from which the selected image was collected and grasped the target object.
The grasp was counted as successful only if the DSR succeeded in grasping the target object.
Finally, the DSR carried the target object to the receptacle, completing the fetch-and-carry task.
This final step was only  attempted
if the users had selected appropriate images for both the target object and the receptacle, and the DSR had successfully grasped the target object.
 The task was considered a success if the DSR successfully carried the target object and placed it on the receptacle. 
We did not utilize any learning-based methods in generating trajectories for object grasping and placing, as this aspect was beyond the scope of our research.

\vspace{-2.7mm}
\subsection{Results}
Table~\ref{tab:quant_physical} presents the quantitative results of the physical experiments for the baseline methods \cite{chen23icra, kaneda24ral, korekata24arxiv} 
(as described in Section~\ref{quantitative_res}) and the proposed methods.
The evaluation metric for the physical experiments was the task success rate (SR), which is a standard evaluation metric for robotic manipulation tasks.
SR is defined as $\text{SR}={{N}_\text{s}}/{{N}_\text{a}}$,
where $N_\text{s}$ and $N_\text{a}$ denote the numbers of successes and attempts, respectively.
The grasping attempts were limited to instances in which the image of the target object was within the top-10 results retrieved in the target mode.
Likewise, placing actions were only attempted when the following two conditions were met: the image of the receptacle ranked among the top-10 in receptacle mode, and the preceding grasping action had been successful.
 \begin{table}[t]
    \vspace{-0.8mm}
    \caption{\small Quantitative comparison between RelaX-Former and baseline methods for physical experiments. The numbers in parentheses show the SR ($N_\text{s}/N_\text{a}$). The best score is in \textbf{bold}.}
    \vspace{-1mm}
    \label{tab:quant_physical}
    \centering
    \normalsize
    \begin{tabular}{@{\hspace{6mm}}c@{\hspace{10mm}}l@{\hspace{10mm}}c@{\hspace{6mm}}}
       \toprule
         & Method &  SR$\uparrow$\,[\%] \\
        \midrule
        (i) & NLMap\cite{chen23icra} (rep.) &  
        64 {(64\,/\,100)}
        \\
        (ii) & MultiRankIt\cite{kaneda24ral}  &
         53 {(53\,/\,100)}
        \\
        (iii) & $\text{DM}^{2}\text{RM}$ \cite{korekata24arxiv} & 
         68 {(68\,/\,100)}
        \\
        \midrule
        (iv) & RelaX-Former (ours) &
         \textbf{75} (75\,/\,100) \\
        \bottomrule
    \end{tabular}
    \vspace{-2.8mm}
\end{table}

Table~\ref{tab:quant_physical} indicates that the proposed method (iv) achieved an overall SR of
75\%, 7
points higher than the best baseline method (iii).
The reason that method (i) was better than method (ii) in the real-world experiment, but worse in the simulated experiments, is because its training-free nature and dependence on CLIP limited its ability to handle the complex referring expressions frequently present in the instructions within the dataset.

\begin{figure}[t]
    \vspace{1.2mm}
    \centering
    \includegraphics[width=\linewidth]{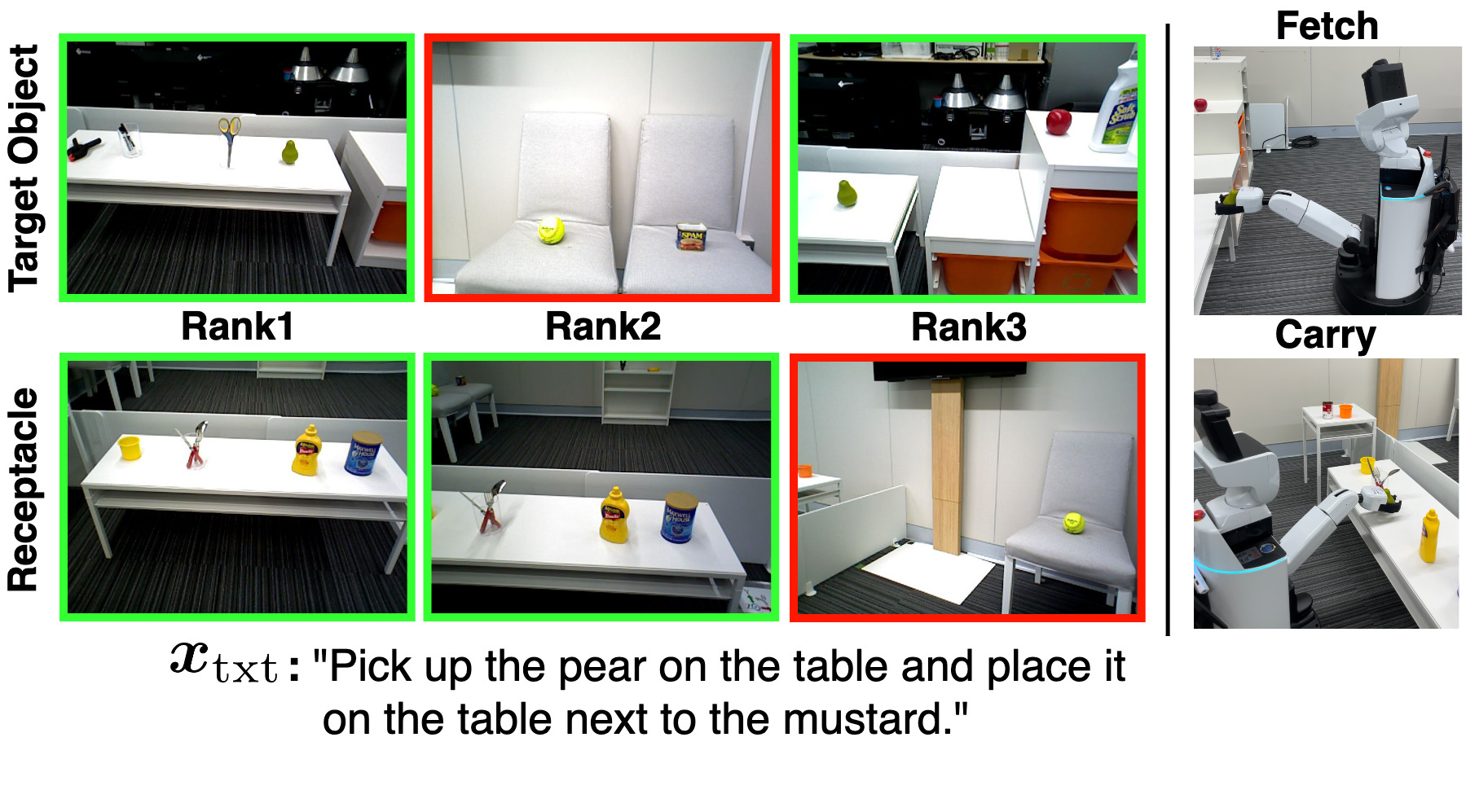}
    \vspace{-9mm}
    \caption{\small Qualitative results of the physical experiments with a given instruction $\bm{x}_{\text{txt}}$. The top-3 retrieved images for each mode are shown, alongside the scenes of fetching and carrying actions. The images that were considered correct or incorrect for each mode are framed in \textcolor[rgb]{0.2,0.8,0}{green} or \textcolor[rgb]{1.0,0,0}{red}, respectively.}
    \label{fig:qualitative_wrs}
    \vspace{-6mm}
\end{figure}

Fig.~\ref{fig:qualitative_wrs} shows a successful example of the physical experiments.
The $\bm{x}_{\text{txt}}$ was ``Pick up the pear on the table and place it on the table next to the mustard.''
In target mode, the proposed method was able to retrieve the correct image in first place.
For receptacle mode, the top-2 images retrieved by the proposed method were correct.
These two images can both be considered correct because they contain the same scene captured from different angles.
Subsequently, the DSR grasped the pear and placed it on the correct table.

\vspace{-2.8mm}
\section{Conclusions}
\vspace{-0.8mm}

In this study, we focused on the IROV-FC task\cite{korekata24arxiv}. 
In this task, the DSR retrieved images of the target object and the receptacle based on an open-vocabulary instruction and subsequently transported the target object to the receptacle.
We proposed RelaX-Former, a method that leveraged unlabeled positive samples and introduced a 
double
relaxed contrastive learning approach to handle the relations among positive, unlabeled positive, and negative samples.
RelaX-Former outperformed the baseline methods in terms of standard metrics on the LTRRIE-FC dataset\cite{korekata24arxiv}.
Furthermore, in physical experiments using a DSR, RelaX-Former demonstrated robust performance in a zero-shot transfer setting, achieving an overall success rate of 
75\%.
\section*{ACKNOWLEDGMENT}
\vspace{-0.9mm}
\vspace{-1.2mm}

This work was partially supported by JSPS KAKENHI Grant Number 23K03478, JST Moonshot, and NEDO.

\bibliographystyle{IEEEtran}
\bibliography{reference}

\end{document}